\documentclass[letterpaper, 10 pt, conference]{ieeeconf}  

\IEEEoverridecommandlockouts                              

\pdfoutput=1

\usepackage[british,ngerman]{babel}
\usepackage[T1]{fontenc}
\usepackage[utf8]{inputenc}
\usepackage{listings} 
\usepackage{graphicx} 
\usepackage[plainpages=false]{hyperref} 

\hypersetup{
	colorlinks,
	linkcolor={black},
	citecolor={black},
	urlcolor={black}
} 

\title{\LARGE \bf
Towards a graphical language for quadrotor missions
}

\author{Benjamin Schwartz\textsuperscript{1}, Ludwig Nägele\textsuperscript{1}, Andreas Angerer\textsuperscript{1}, and Bruce A. MacDonald\textsuperscript{2}
\thanks{\textsuperscript{1} Institute of Software \& Systems Engineering, University of Augsburg, Germany}%
\thanks{\textsuperscript{2} CARES, Department of Electrical and Computer Engineering, University of Auckland, New Zealand}
}

\begin{document}

\setlength{\dbltextfloatsep }{.5\baselineskip} 

\maketitle
\thispagestyle{empty}
\pagestyle{empty}
\selectlanguage{british}

\begin{abstract}
This paper presents an approach for defining Unmanned Aerial Vehicle (UAV) missions on a high level. Current methods for UAV mission specification are evaluated and their deficiencies are analyzed. From these findings, a new graphical specification language for UAV missions is proposed, which is targeted towards typical UAV users from various domains rather than computer science experts. The research is ongoing, but a first prototype is presented.
\end{abstract}


\section{Introduction}
Quadrotors or in general Unmanned Aerial Vehicles (UAV) are becoming increasingly popular and affordable for a broad range of activities. Researchers and developers are exploring the use of UAVs in application domains such as agriculture, archaeology or infrastructure monitoring. Patel et al. investigate how to detect possible crop diseases on fields using a quadrotor equipped with an infrared camera~\cite{Patel}. Employing  quadrotors for overhead pictures of an exhibition site to reduce the cost for archaeological research is discussed by Heiermann et al.~\cite{Heiermann}. The German companies Thyssengas and microdrones are working on a system to inspect gas pipelines with quadrotors~\cite{Thyssengas}. The plans of the company Amazon to use UAVs to deliver packages~\cite{Amazon} to end customers went through the international press some months ago. 

To execute such tasks reliably, UAVs must be increasingly equipped with more autonomy. The routes the UAV should take might be predefined or may depend on parameters of the task, additional actions might be necessary at certain points of the route (such as taking pictures, activating sensors) and unforeseen situations must be handled (e.g. avoiding obstacles). Above all this, the operational state, including the battery lifetime, and environment conditions, such as wind and weather conditions, should be checked constantly to safely abort the mission if necessary.

The predominant way of defining and creating such a UAV mission is hard-coding it in a general-purpose programming language. However, the typical users of UAV systems -- e.g., farmers, archeologists or pipeline technicians -- do not have the skill to design software of this complexity by hand. This paper presents a novel approach to describe UAV missions with a graphical Domain Specific Language based on an analysis of use cases and grouping of concepts, and with an extendable model. Sect.~\ref{sec:analysis} presents existing approaches and motivates the decision to design a new domain specific language (DSL). Concepts and challenges are introduced in Sect.~\ref{sec:concept}. The work is ongoing, however, the current prototype and first results are presented in Sect.~\ref{sec:prototype}.

\section{Analysis} \label{sec:analysis}
The first part of this section examines existing tools for defining UAV missions in a user-friendly way and motivates the choice to create a new language. In the second part, different approaches for realizing such a language are discussed.


\subsection{Related work}



MissionLab developed in '99 by Arkin et al.~\cite{Arkin1999a} allows specifying missions for autonomous robots in a graphical way. The interface, however, is quite restricted. In '04 a mission planning wizard was added to improve usability. A usability evaluation~\cite{Endo2004a} showed it improved and sped the mission planning process. However, MissonLab lacks flexibility for more complex, branched workflows.

MAVLink based robots use the APM Planner~\cite{ArdupilotMegaProject} mission planning tool, which has a Graphical User Interface (GUI) to configure both waypoints and tasks to be carried out at each waypoint. This tool does not support conditional waypoints or specifying simultaneous tasks such as avoiding obstacles. The APM Planner supports communication with a ground station during the flight, which enables manual reactions by the operator of the station to specific changes. This communication is mandatory and must be maintained during the whole flight, which limits missions to a specific area. The GUI is highly developed and can also show the waypoints of a mission in a virtual map. However, the planned mission cannot be changed after generation and cannot be adapted to the user needs. DJI~\cite{DJI} provides a similar proprietary tool for their quadrotors. An additional feature of this tool is the ability to define non-fly zones.

Graphical State Space Programming (GSSP)~\cite{Li2011c} is another framework for developing quadrotor missions. GSSP has a graphical and a textual interface. The graphical part is used to declare waypoints and the workflow of the mission. Special tasks, which are carried out during the execution, are specified in the textual interface in the programming language python. Thus, it is possible to perform operations during the flight simultaneously or introduce branches in the workflow. However, even the simplest of such special tasks requires writing code.

The analysis showed some major deficiencies. Many existing approaches (e.g. APM Planner and DJI PC Ground Station) have a proprietary character and can only be used with certain types of UAVs. Moreover, though providing graphical interfaces, all approaches focus on the main, sequential workflow of a UAV mission and are very limited when it comes to branches and alternative flows. In some approaches, additional flexibility is achievable by adding new functionality using some host programming language. This is however not optimal for end users that lack computer science background. Finally, similar deficiencies exist with respect to simultaneous tasks that have to be executed during the whole mission or parts of it. Existing approaches provide little to no graphical support for this.

To overcome the identified problems and provide a flexible and at the same time easy-to-use description mechanism, the authors decided to design a new graphical language that should incorporate advantages of existing work, but add much more flexibility for describing UAV missions while still being intuitive to use also for non-experts. 

\subsection{Graphical programming languages}
When creating a graphical language, a straightforward approach is to use an existing general-purpose visual programming tool.
A prominent example is Scratch~\cite{Resnick2009}, which is used to help people learn to program textual general-purpose programming languages. It provides a drag and drop mechanism to put together an application. Since its purpose is to help people learn a general-purpose programming language it focuses on traditional programming paradigms. A related project is MIT's App Inventor~\cite{Wolber2011}, which uses Scratch for rapidly prototyping Android applications without programming knowledge. Another powerful general-purpose visual programming tool is Microsoft Visual Programming Language (VPL)~\cite{Microsoft}, which employs a dataflow-based programming model. Developing a 'new' language based on Scratch or VPL would be possible with moderate effort. However, some advanced domain specific concepts would likely require modifications to be made in the respective editor, which is not possible since VPL is a closed source. Otherwise, all domain-specific concepts would need to be encoded based on existing syntactic elements.

A second approach is to employ a toolkit for constructing a new DSL.
The advantages of a custom DSL are the higher abstraction level, which enables domain experts to create programs in terms of familiar domain concepts. A DSL can furthermore provide domain specific validations.
A drawback is the effort to create a DSL, which only pays off if the DSL can be used frequently. Often different DSLs exist in one domain and there is seldom a standard DSL, which can lead to confusion for the user. An executable DSL can also produce code, which can be used as a starting point for a quadrotor mission. Existing frameworks to support the development of a DSL include the Eclipse Modeling Framework~\cite{Gronback2009,TheEclipseFoundation}.

Even more freedom is achievable with a new custom designed editor implemented in a general-purpose language. APM Planner and PC Ground Station of DJI are examples. The advantage is the broad range of possible designs.
A drawback of this freedom is that the resulting tool will be more complicated for other developers to extend and enhance, as well as development time and cost will be higher than with both other methods.

The DSL toolkit approach is more flexible and allows good tailoring of the developed language to the specific needs of the domain.
\begin{figure*}[!ht]
	\centering
	\includegraphics[width=0.97\linewidth]{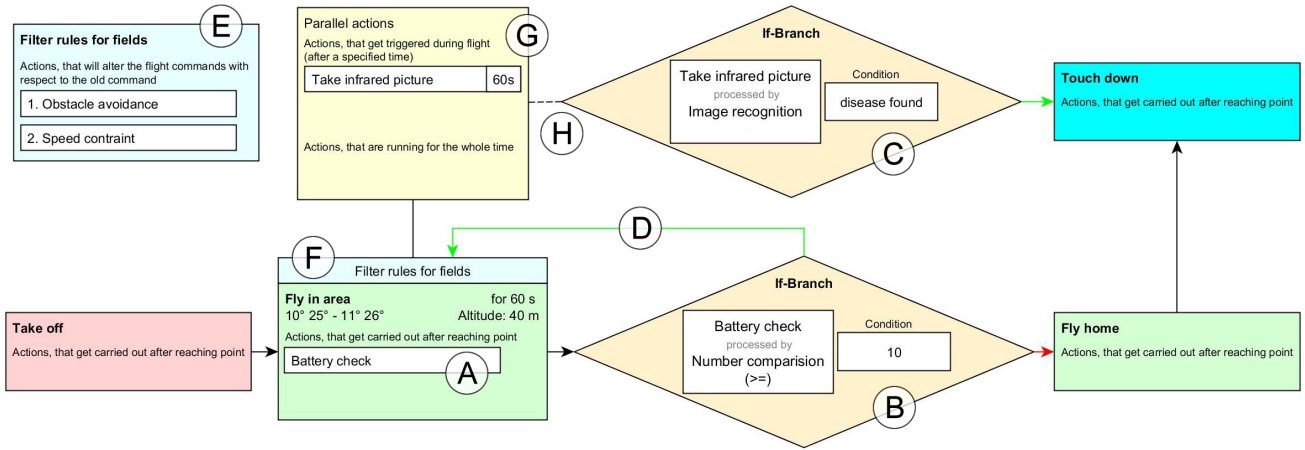}
	\caption{Example quadrotor mission program which demonstrates the invented language constructs}
	\label{fig:Editor2}
\end{figure*}

\section{Proposed graphical language} \label{sec:concept}
The conceptual model of the developed language is based on a literature research, identifying numerous use cases whose purposes vary significantly. A common set of concepts was identified and used as basis for the language. The first part of this section will introduce those concepts, whereas the second part will present the graphical syntax of the proposed language.

\subsection{Basic concepts}


\textbf{Routing elements} are the central concept to describe quadrotor missions. They are used to move the quadrotor from some location to another one. Examples for routing elements are taking off, touching down or flying to a certain position or area.

While routing elements provide means for building the structure of a mission, \textbf{actions} can be used to give a mission a purpose. In the analyzed use cases, a broad range of activities was carried out at various points during a mission. Examples include taking a picture, using a laser scanner, or scanning for a signal such as wireless LAN. The idea of actions is similar as in GSSP. However, in contrast to GSSP the implementation is always hidden from the user. 

To introduce flexibility in the mission flow, \textbf{conditional branches} are available. They allow for specifying alternative mission flows depending on runtime conditions. This concept is adapted from general programming language theory and allows expressing a broad range of quadrotor missions.

For conditional branching, input data is needed as well as some kind of processing of this data. This led to the introduction of \textbf{processing actions}, e.g. to recognize an image or interpret a laser scan.

For enhancing routing elements with individual movement strategies, the concept of \textbf{filters} is introduced. This is based on the observation that quadrotor movements frequently need to respect certain constraints, e.g. maintaining a specified velocity or avoiding obstacles on-line. Thus, filters may influence routing elements.

Finally, to support activities that are meant to run simultaneously to the main flow, a \textbf{parallel} block is provided. This can be employed e.g. to monitor sensor outputs, record a video or take a picture every minute. In conjunction with conditional branches, parallel blocks are able to influence the program flow, e.g. if a sensor value exceeds some threshold.

\subsection{Graphical syntax}
To gain a consistent visual language all concepts mentioned above and the way they interact need to obtain meaningful graphical representations. In order to keep the language clear and intuitive, different colors are used to easily distinguish between different element types. 

The routing elements \textit{Take off} and \textit{Touch down} are the starting and ending points of every mission and thus form an invariant surrounding of each program.
Within this surrounding, a chronological order of routing elements (e.g. \textit{Fly home}, \textit{Fly in area}, ...) is possible (cf. Fig.~\ref{fig:Editor2}).
As discussed above, routing elements should be augmentable with actions that are to be performed at various points of the mission. 
Instead of connecting actions to (possibly many) routing elements by using (many) links, the design of our language embeds instances of actions directly in routing elements (Fig.~\ref{fig:Editor2}, A).

Conditional branches are symbolised by diamonds (Fig.~\ref{fig:Editor2}, B) having two outgoing transitions, a green one to be used in case of true and a red one for false.
For validation purpose, conditional branches reference the result of one previously executed action (of the previous block or element), processes it in some way (using processing actions) and compares the output to the given reference value. In the given example (Fig.~\ref{fig:Editor2}, C), the infrared picture taken in the previous parallel block is processed by an image recognition action, which may produce the result ``disease found''.
Also cycles and iterations can be modeled with conditional branches (Fig.~\ref{fig:Editor2}, D).
Filters have to be defined by specifying a list of atomic filter actions. The order of those actions specifies their priority.
In the example (Fig.~\ref{fig:Editor2}, E) the filter monitors a certain speed value but always guarantees collision freedom. Once defined, a filter can be attached to routing elements (Fig.~\ref{fig:Editor2}, F).

Parallel blocks (Fig.~\ref{fig:Editor2}, G) can be attached via links to one or more routing elements to which the embedded actions should run concurrently. For periodically triggered actions, a period time can be specified. As parallel blocks should be able to alter the mission workflow in some cases, conditional branches can be attached to them. However, in order to preserve a clean main mission flow, this combination implies a particular semantics: If a parallel block is followed by a conditional branch, the false-case implicitly returns control to the parallel block. Thus, the execution context leaves the parallel block only in the true case. To illustrate this special semantics, a dashed line to the conditional branch is used instead of an arrow (Fig.~\ref{fig:Editor2}, H). In case of evaluation to true, the green arrow indicates the next active routing element of the main flow.

\begin{figure*}[!ht]
	\centering
	\includegraphics[width=0.97\linewidth]{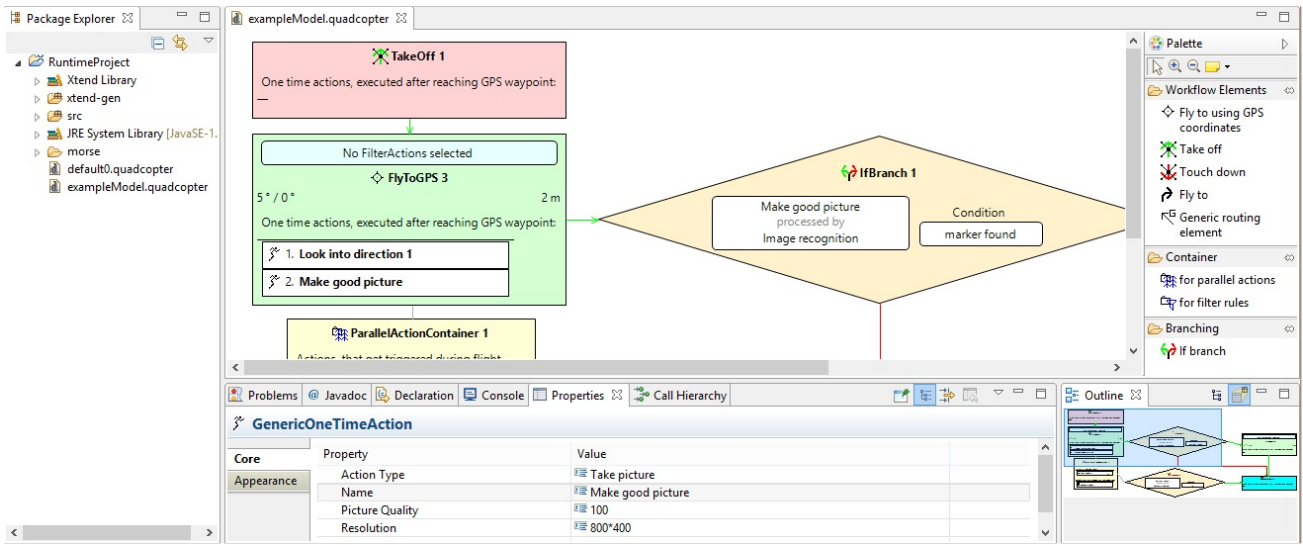}
	\caption{Example of the invented language in the developed editor}
	\label{fig:Editor}
\end{figure*}

\section{Prototype} \label{sec:prototype}
A prototype (see Fig.~\ref{fig:Editor}) has been implemented based on the Eclipse Modeling Project (EMP)~\cite{Gronback2009}, which is a set of different model-based development tools and frameworks.
A meta-model of the designed quadcopter mission language has been defined and a graphical editor could be generated by using EMP's Graphical Modelling Framework (GMF).
The extension by new actions was identified as an important feature of the language, because not all possibly desired functionality of quadrotor missions can be predicted beforehand and considered in the meta-model.
Since it would be highly impractical to adapt the meta-model and recompile the editor for each new action, another way of enhancing the editor by additional actions has been developed.
In the following first section the method how to add new concrete action implementations is presented. A brief look onto the execution of developed quadrotor programs is given in the subsequent section.


\subsection{Extensions mechanism for user-specific actions}

For extending the program with own actions, a super class \emph{GenericAction} is provided. It is part of the meta-model and needs to be inherited programmatically by all action implementations by declaring special information such as a name for describing the action (e.g. ``TakePicture''), the return type of the action and its logical behaviour.
An action implementation can then graphically be used - maybe more than once - within the graphical program in the editor.
For each usage, a new instance of the action type is created. In case the action is capable of being parameterised, the super class ensures that each instance can be given individual specific configurations. Technically, this is a map of variable name and value pairs. For the ``TakePicture''-example, additional parameters might be resolution and compression quality.
The properties view (see Fig.~\ref{fig:Editor}, bottom middle) shows these variables and allows for their modification. The possibility of parameterising makes actions reusable. The extension by new processing actions works analogously.
A specific action can be implemented, graphically referenced and parameterised within the same single development environment.

Regarding different result types of action definitions, the prototype ensures type consistency when actions are to be referenced in processing actions or comparisons within conditional branches.

\subsection{Execution}
Different techniques to execute the model were considered.
In general interpreting code is less efficient than executing previously generated code. Furthermore, an interpreter typically needs more hardware resources which might not be available on limited quadrotor hardware. Thus, code generation was chosen for the prototype.
In order to have a first evaluation of the invented language using a real-world application and to be able to rapidly prototype and test graphical programs, Morse~\cite{Echeverria2011} and its quadrotor extension is used as standalone application. Python code, which controls the Morse simulator, is generated out of the editor by using Xtend~\cite{bettini2013implementing} template expressions.

\section{Conclusion \& Future work} \label{sec:conclusion}
The paper showed there is need for a higher-level and platform-independent approach for specifying quadrotor missions. A graphical DSL was identified to have the highest potential to solve the given requirements and issues, and a set of concepts was discussed to represent the structure of the new language. The implemented prototype showed that the identified theoretical domain concepts can be automatically translated into working applications.

A future step is to conduct a user-study to see how the chosen graphical representation and the editor are used and to identify possible improvements in order to simplify usage. Such an improvement would be a virtual map for showing waypoints. Another further development could be the extension by new translation templates for supporting code translation to real quadrotor platforms. The model could also be enriched by more environment information such as where the quadrotor is flying. This could result in a different collection of rules, for example touch-down is not allowed over water.

\bibliographystyle{IEEEtran_custom}
\bibliography{IEEEabrv,Paper}
\end{document}